# Computational Imaging for Enhanced Computer Vision

Humera Shaikh, Kaur Jashanpreet
Faculty of Engineering and Technology
SRM Institute of Science and Technology, Chennai

*Abstract*—This paper presents a comprehensive survey of computational imaging (CI) techniques and their transformative impact on computer vision (CV) applications. Conventional imaging methods often fail to deliver high-fidelity visual data in challenging conditions, such as low light, motion blur, or high dynamic range scenes, thereby limiting the performance of state-of-the-art CV systems. Computational imaging techniques, including light field imaging, high dynamic range (HDR) imaging, deblurring, high-speed imaging, and glare mitigation, address these limitations by enhancing image acquisition and reconstruction processes. This survey systematically explores the synergies between CI techniques and core CV tasks, including object detection, depth estimation, optical flow, face recognition, and keypoint detection. By analyzing the relationships between CI methods and their practical contributions to CV applications, this work highlights emerging opportunities, challenges, and future research directions. We emphasize the potential for task-specific, adaptive imaging pipelines that improve robustness, accuracy, and efficiency in real-world scenarios, such as autonomous navigation, surveillance, augmented reality, and robotics.

## I. INTRODUCTION

COMPUTER vision (CV) is a rapidly evolving discipline within artificial intelligence (AI) that enables machines to analyze, interpret, and understand visual data from the real world [1]. By processing digital images and videos, CV sys- tems accomplish a variety of tasks, including object detection [2], [3], face recognition [4]–[6], depth estimation [7], [8], keypoint detection [9], [10], and optical flow estimation [11], [12]. These tasks are critical for applications ranging from autonomous vehicles and robotics to medical imaging and augmented reality. Despite the remarkable progress in recent years, the performance of CV systems remains constrained by the quality of input images, which often fail to represent the environment fully and accurately. However, in the age of machine learning [13] in general and deep learning in particular in various imaging applications the inadequacy of conventional images is felt more than ever before.
Standard imaging techniques suffer from limitations such as motion blur, low dynamic range, insufficient depth informa- tion, and glare, which reduce the fidelity and richness of the captured data. These shortcomings become particularly pronounced in challenging conditions, such as low light, high-speed motion, or extreme contrast, where traditional imaging struggles to provide reliable inputs for CV algorithms. To overcome these limitations, computational imaging (CI) has emerged as a groundbreaking approach that integrates optical design with computational algorithms to capture, process, and reconstruct richer and more informative representations of a scene [14]. Unlike conventional imaging systems that merely record 2D intensity information, computational imag- ing techniques exploit additional dimensions, including spatial, angular, temporal, and spectral data, to better represent the complexities of real-world environments.

Techniques such as light field imaging [15], [16] provide multi-perspective views of a scene for enhanced depth esti- mation, while high dynamic range (HDR) imaging [17]-[19] captures details in both bright and dark regions, overcoming the constraints of conventional sensors. Similarly, methods like deblurring [20]-[22] and glare mitigation [23], [24] address issues caused by motion or unfavorable lighting, and high- speed imaging [25], [26] enables the analysis of fast-moving objects with unprecedented clarity. By producing high-quality data enriched with additional information, CI techniques pro-vide more reliable inputs for CV applications, significantly improving their accuracy and robustness. This survey is dedicated to exploring how computational imag- ing techniques enhance computer vision applications, address- ing gaps in existing research where the two domains intersect. While previous works have discussed computational imaging methods in isolation or focused on individual computer vision tasks, this survey takes a comprehensive and unified approach. It systematically examines a wide range of computational imaging techniques --- light field imaging, HDR imaging, de-blurring, high-speed imaging, and glare mitigation-and ana- lyzes their impact on multiple computer vision tasks, including object detection, face recognition, depth estimation, keypoint detection, and optical flow estimation. By bridging the gap be- tween computational imaging and computer vision, this work highlights the synergies that emerge when advanced imaging techniques are combined with state-of-the-art (SOTA) CV methods. Specifically, this survey contributes to the literature in several ways. First, it provides an in-depth explanation of major computational imaging techniques, including their prin- ciples, methodologies, and underlying mathematical models, offering readers a clear understanding of how these techniques improve image acquisition and processing.

IJERTV14IS050351

Second, it highlights the specific computer vision tasks that benefit from these techniques, detailing how CI methods address existing limitations in SOTA CV algorithms.

For instance, light field imaging significantly enhances depth esti- mation and keypoint detection by capturing angular and spatial details, while HDR imaging improves object detection and face recognition in high-contrast environments. Third, this survey is unique in its breadth, covering a large variety of computational imaging methods and computer vision applications within a single work, making it a valuable resource for researchers and practitioners working at the intersection of these fields. The organization of the paper reflects this comprehensive scope. Section 2 provides a detailed overview of computational imaging techniques, explaining their mathematical foundations and methodologies. Section 3 explores key computer vision applications, discusses SOTA methods for each task, and highlights how computational imaging techniques enhance their performance. Section 4 summarizes the relationships between CI techniques and CV applications, offering insights into their practical synergies and use cases. Section 5 of the paper discusses future directions, challenges, and opportunities for integrating computational imaging and computer vision. Finally, Section 6 concludes the paper.

Through this survey, we aim to emphasize the transformative potential of computational imaging techniques in advancing computer vision, while offering a unique perspective that unifies these two domains. By shedding light on the role of CI in improving the quality and reliability of visual data, this work serves as a foundation for future research and development at the intersection of computational imaging and computer vision.

## II. COMPUTATIONAL IMAGING TECHNIQUES

Computational imaging (CI) integrates advanced optical hardware with computational algorithms to overcome the limitations of traditional imaging systems. By capturing richer, multidimensional information such as angular, spatial, spectral, and temporal details [27], CI techniques allow the reconstruction of more accurate representations of a scene. This section provides an in-depth explanation of key computational imaging methods, including their principles, methodologies, and impact on computer vision tasks.

### A. Light Field Imaging

Light Field Imaging [28], [29] is a transformative technique that captures not only the intensity of light but also its angular direction. Unlike conventional cameras, which record a 2D projection of the 3D scene, light field cameras equipped with micro-lens arrays (MLAs) sample the 4D light field function $L(u, v, s, t)$, can be represented as follows:

$$L(u, v, s, t) = \int \delta(u - x, v - y) I(x, y, s, t) \, dx \, dy,$$

where $(u, v)$ represent the spatial coordinates on the image plane, and $(s, t)$ represent the angular coordinates indicating the direction of light rays. By sampling light rays from multiple perspectives within a single exposure, light field cameras provide rich angular diversity that enables post-capture operations such as refocusing, depth estimation, and occlusion handling.

Depth estimation is one of the primary applications of Light Field Imaging, as angular variations allow precise reconstruction of the depth map. The disparity $d$ between views in the light field is directly related to the depth $Z$ of a point in the scene, given by:

$$Z = \frac{f \cdot \Delta u}{d}, \quad (1)$$

where $f$ is the focal length of the imaging system, $\Delta u$ is the baseline or separation between angular views, and $d$ is the measured disparity. This relationship allows algorithms to compute dense depth maps by analyzing shifts in angular views, making light field imaging invaluable for computer vision tasks such as *3D reconstruction* and *face recogination* [30], [31].

Beyond depth estimation, Light Field Imaging supports refocusing at different planes in the scene, a feature particularly beneficial for applications involving objects at varying depths. This property has found significant utility in autonomous systems [32], where precise depth perception and occlusion reasoning are critical for navigation and obstacle avoidance. Moreover, Light Field Imaging enhances augmented and virtual reality [33], [34] applications by allowing real-time synthesis of new viewpoints, ensuring seamless visual immersion. The angular richness provided by light field cameras has also been leveraged for improving object segmentation and scene understanding, as the additional perspectives allow better differentiation between objects and their backgrounds.

### B. High Dynamic Range Imaging

High Dynamic Range (HDR) Imaging [35], [36] is designed to address the limitations of traditional sensors, which struggle to capture details in both bright and dark regions of a scene. HDR techniques overcome this issue by capturing multiple images of the same scene at different exposure levels and fusing them into a single image with an expanded dynamic range. Mathematically, the HDR image can be computed as:

$$I^{HDR}(x, y) = \frac{\sum_{i=1}^{N} w(E_i(x, y)) E_i(x, y)}{\sum_{i=1}^{N} w(E_i(x, y))}, \quad (2)$$

where $E_i(x, y)$ is the radiance at pixel $(x, y)$ from the $i$-th exposure, and $w$ is a weighting function that reduces the influence of saturated or underexposed pixels.

HDR imaging is particularly beneficial for computer vision tasks operating in high-contrast environments, such as outdoor object detection [37], [38] and face recognition under varying illumination. For instance, in autonomous driving systems [39], HDR images ensure visibility of road details and traffic signals that are either backlit or illuminated by strong headlights. Similarly, in surveillance and robotics [40], HDR imaging provides reliable inputs by preserving details across shadowed and brightly lit regions, significantly enhancing the performance of feature extraction [41], segmentation [42], and image classification algorithms [43].

C. Image Deblurring

Image Deblurring is a key computational imaging method that mitigates the degradation caused by motion blur and defocus blur [44], [45]. Motion blur occurs when the camera or scene moves during exposure, while defocus blur arises due to a limited depth of field that fails to keep all regions of the scene in focus. Deblurring techniques model the imaging process as a convolution operation, where the observed blurred image $I_{blurred}$ is expressed as:

$$I_{blurred} = I_{sharp} \otimes k + n, \qquad (3)$$

where $I_{sharp}$ is the original sharp image, $k$ is the blur kernel (or point spread function), $\otimes$ represents the convolution operation, and $n$ is additive noise. Blind deblurring methods aim to estimate both the sharp image $I_{sharp}$ and the unknown blur kernel $k$ using iterative optimization or learning-based approaches. These techniques play a critical role in enhancing computer vision tasks such as optical flow estimation, motion tracking, and object detection, where sharp edges and well-defined textures are essential for accurate feature extraction. For example, in surveillance and robotics, deblurring restores image clarity in dynamic environments, enabling robust detection and tracking of fast-moving objects.

D. High-Speed Imaging

High-Speed Imaging focuses on capturing rapid temporal events that cannot be resolved with conventional cameras operating at standard frame rates [25], [46]. By increasing the temporal resolution, high-speed imaging systems record fine-grained motion details, enabling the analysis of dynamic scenes such as fast-moving objects, rapid biological processes, and industrial inspection tasks. Event-based cameras are particularly notable in this domain, as they record pixel changes asynchronously, reducing data redundancy and achieving ultra-fast motion capture.

The ability to resolve high-speed motion without introducing motion blur significantly benefits computer vision tasks such as object tracking, optical flow computation, and motion analysis. For instance, in industrial settings, high-speed imaging enables defect detection on high-speed production lines, while in sports analysis, it provides precise tracking of equipment and player movements.

E. Glare Mitigation

Glare Mitigation addresses artifacts caused by reflections and strong directional light sources, which obscure critical scene details [47], [48]. Techniques such as polarization-based imaging filter light waves at specific orientations to suppress reflections, while multi-exposure methods capture images under varying lighting conditions and computationally combine them to eliminate glare. These approaches are particularly beneficial for applications like object detection and face recognition in reflective environments, including glass surfaces, water bodies, and metallic objects. By reducing glare, these techniques ensure that computer vision systems maintain high accuracy in challenging lighting conditions.

Computational imaging techniques such as light field imaging, HDR imaging, deblurring, high-speed imaging, and glare mitigation enable the acquisition of richer and more accurate visual data. These techniques address long-standing challenges such as limited dynamic range, motion blur, glare, and fast motion, providing high-quality inputs for computer vision systems. By leveraging these methods, CV tasks such as depth estimation, object detection, optical flow, and motion analysis achieve enhanced accuracy and robustness in diverse and challenging scenarios.

## III. COMPUTER VISION APPLICATIONS

Computer vision (CV) encompasses a variety of tasks that enable machines to perceive, analyze, and understand visual data [49]. These tasks play a pivotal role in numerous real-world applications, including autonomous vehicles, surveillance, augmented reality, robotics, agriculture, and medical imaging. However, CV systems are often constrained by the quality of input data, which can be degraded due to low lighting, motion blur, occlusions, or limited dynamic range. Computational imaging (CI) techniques provide innovative solutions to overcome these challenges, thereby enhancing the robustness, accuracy, and efficiency of CV systems. This section discusses key computer vision applications—object detection, depth estimation, optical flow, and keypoint detection—and explains how CI techniques contribute to their improvement.

### A. Object Detection

Object detection is one of the most fundamental tasks in computer vision, where the goal is to identify and locate objects within an image or video frame [50]. It serves as the backbone for applications such as autonomous vehicles (AVs) [51]–[53], drone surveillance [54], precision agriculture [55], [56], and industrial inspection [57]. Mathematically, object detection involves mapping the input image $I$ to a set of bounding boxes and corresponding class probabilities:

$$D(I) = \{(b_i, p_i) \mid i = 1, \ldots, N\}, \quad (4)$$

where $b_i$ represents the bounding box for the $i$-th detected object, $p_i$ denotes the probability of the object belonging to a certain class, and $N$ is the total number of detected objects. In real-world scenarios, poor lighting, motion blur, and high dynamic range (HDR) [58] scenes can significantly hinder the performance of object detection algorithms. For instance, in autonomous vehicles operating at night, standard cameras fail to capture sufficient details of pedestrians or vehicles. *HDR imaging* addresses this limitation by combining multiple exposures into a single image that preserves details across both shadows and highlights. This ensures that critical objects, such as traffic signs and obstacles, remain visible even under challenging lighting conditions.

Similarly, *image deblurring* improves object detection in dynamic environments where motion blur is prevalent. For example, surveillance cameras tracking fast-moving individuals or drones monitoring agricultural fields often produce blurred frames, which obscure object boundaries. Deblurring algorithms restore these frames by solving for the sharp image using the blur kernel, thus improving the accuracy of object localization. Additionally, light field imaging helps resolve occlusions in crowded scenes, such as those encountered in drone-based surveillance or industrial monitoring, by capturing angular information that separates objects from their backgrounds.

### B. Depth Estimation

Depth estimation is a critical CV task that involves inferring the 3D structure of a scene from 2D images. It is essential for applications like robotic navigation, augmented reality (AR), virtual reality (VR), and autonomous systems. The relationship between the disparity $d$, baseline $\Delta u$, and depth $Z$ is described as follows:

$$Z = \frac{f \cdot \Delta u}{d}, \quad (5)$$

where $f$ is the focal length of the imaging system, $\Delta u$ is the distance between viewpoints (baseline), and $d$ represents the disparity between corresponding points in two or more images. Conventional stereo vision systems rely on a limited number of viewpoints, which can lead to inaccuracies in the presence of occlusions or textureless regions. *Light field imaging* overcomes these challenges by capturing the 4D light field function $L(u, v, s, t)$, which contains both spatial and angular information [59]–[61]. By analyzing angular shifts across multiple viewpoints, light field techniques produce dense and accurate depth maps [62], [63] that are particularly useful for autonomous vehicles navigating cluttered environments, where precise depth information is crucial for obstacle avoidance.

In AR and VR applications, accurate depth estimation allows virtual objects to seamlessly integrate into real-world scenes, enhancing user immersion. Additionally, agricultural robotics use depth estimation to detect fruits, plants, and obstacles, facilitating automated harvesting and navigation. In medical imaging, depth-aware techniques improve the visu- alization of anatomical structures during minimally invasive surgeries, ensuring better precision and safety.

### C. Optical Flow for Tracking

Optical flow estimation is used to compute the motion of pixels between consecutive frames in a video sequence. It plays a key role in applications such as object tracking, motion analysis, video stabilization, and autonomous navigation. The motion field $V(x, y, t) = (u, v)$ describes the horizontal ($u$) and vertical ($v$) flow components at a given pixel location ($x, y$) and time $t$. The optical flow constraint assumes that the brightness of pixels remains constant between frames, as expressed by:

$$I(x, y, t) = I(x + u, y + v, t + 1). \quad (6)$$

In practice, optical flow estimation is highly sensitive to motion blur and low frame rates, especially in scenes involving fast-moving objects. *High-speed imaging* mitigates these challenges by increasing the temporal resolution, capturing fine-grained motion details that prevent pixel displacement errors. For example, in drone-based wildlife monitoring, high-speed imaging enables accurate tracking of fast-moving animals, even in dense and dynamic environments.

Additionally, *image deblurring* complements optical flow methods by restoring sharp edges and textures, which are critical for accurate motion estimation. This is particularly valuable for surveillance systems and autonomous vehicles, where motion blur can obscure object trajectories and hinder real-time decision-making.

### D. Keypoint Detection and Matching

Keypoint detection involves identifying prominent features, such as corners, edges, or textures, within an image. These features are then used for tasks like image stitching, 3D reconstruction, and simultaneous localization and mapping (SLAM). Keypoint detectors, such as SIFT (Scale-Invariant Feature Transform) and ORB (Oriented FAST and Rotated BRIEF), extract features that are matched across images to estimate scene correspondences.

Keypoint detection often struggles in low-light environments, high-contrast scenes, and occlusions. *Light field imaging* improves keypoint detection by leveraging angular information to resolve ambiguities and occlusions, which are common in cluttered or partially obscured scenes. For example, in robotic navigation, light field data enhances feature detection and matching accuracy, enabling reliable SLAM in indoor and outdoor environments.

*HDR imaging* further enhances keypoint detection in high-contrast scenes, such as urban settings with bright sunlight and deep shadows. By preserving details across varying illumination levels, HDR techniques ensure that keypoints remain consistent and detectable. In AR and VR, this improves the robustness of feature matching, allowing virtual objects to align seamlessly with real-world scenes.

### E. Face Recognition

Face recognition is a computer vision task that involves detecting and identifying human faces in images or video sequences. It is widely used in surveillance systems, access control, biometric authentication, and social media applications. Mathematically, face recognition maps an input face image $I$ to a feature representation $F$ and compares it against a set of stored representations for identification or verification:

$$\text{Similarity}(F_{\text{input}}, F_{\text{stored}}) \geq \tau, \tag{7}$$

where $\tau$ is a predefined threshold, and $F$ represents the extracted feature vector.

Face recognition systems often encounter challenges in real-world scenarios due to poor lighting, glare, occlusions, and variations in facial poses. *Light Field Imaging* addresses these issues by capturing spatial and angular information, allowing the reconstruction of multi-perspective views of the face. By analyzing angular disparities, light field techniques produce a depth-aware representation of the scene, which improves robustness to occlusions and varying facial orientations. This capability ensures reliable face detection and recognition even when parts of the face are obstructed or viewed at extreme angles.

Additionally, light field data enables refocusing at different depths, allowing the face to be isolated from cluttered or blurred backgrounds. For instance, in surveillance applica- tions, where facial recognition often occurs at a distance or in crowded settings, light field cameras enhance detection accuracy by reducing the impact of occlusions and improving feature extraction. Glare caused by reflective surfaces or uneven lighting is also mitigated through light field methods, which can computationally suppress artifacts and enhance the visibility of facial features.

In biometric authentication systems, where high precision is required, light field imaging ensures the robustness of recogni- tion systems by providing additional depth and angular cues. Combined with high dynamic range (HDR) imaging, which preserves facial details under extreme lighting conditions, these CI techniques significantly improve the reliability of face recognition in both indoor and outdoor environments.

## IV. RELATIONSHIPS BETWEEN COMPUTATIONAL IMAGING TECHNIQUES AND COMPUTER VISION APPLICATIONS

The integration of computational imaging (CI) techniques with computer vision (CV) systems has led to transformative advances, not just in individual tasks but also in how these tasks interact and evolve as part of a larger ecosystem. While Section 3 presented specific contributions of CI methods to various CV tasks, this section explores the broader relation- ships, emerging trends, and interdependencies that reveal the full potential of combining these two domains. Computational imaging enables cross-task improvements, where advancements in one CV task propagate to others. For instance, light field imaging not only enhances depth estimation but also indirectly improves downstream tasks such as object detection and keypoint matching. The richer 3D information provided by light field techniques allows object detection algorithms to differentiate overlapping objects in complex environments, while depth-aware keypoints are more robust to occlusions. This synergy is particularly valuable in multi-task systems, such as autonomous navigation, where depth estimation, object detection, and motion tracking are tightly coupled.

Furthermore, CI techniques mitigate long-standing trade-offs in traditional imaging systems. For example, high-speed imaging traditionally sacrifices spatial resolution for temporal clarity, while HDR imaging often struggles with motion artifacts when combining multiple exposures. Modern computational imaging algorithms, however, are increasingly designed to balance these trade-offs. Techniques like event-based cameras in high-speed imaging or deep-learning-based HDR fusion are capable of preserving spatial, temporal, and tonal details simultaneously. These advancements allow CV tasks, such as optical flow estimation and object detection, to operate reliably under previously prohibitive conditions, such as fast motion and extreme lighting.

The integration of CI methods fosters generalizability across environments, addressing the challenges posed by domain shifts. Many CV applications, such as face recognition or surveillance, suffer from performance drops when deployed in varying lighting, angles, or motion conditions. Techniques like HDR imaging and light field imaging provide data that better generalizes across these conditions, reducing the reliance on domain-specific retraining. For example, a face recognition system enhanced with light field imaging can maintain consistent accuracy whether a subject is in low-light conditions, partially occluded, or viewed at an angle.

The synergy between computational imaging and CV also presents new opportunities for real-time processing and deployment. Historically, CI techniques, such as multi-exposure HDR or depth reconstruction from light fields, were computationally expensive, limiting their use in time-critical applications like autonomous driving. However, recent advances in hardware (e.g., specialized sensors, GPUs) and algorithms (e.g., lightweight neural networks) are enabling the real-time deployment of CI-enhanced CV systems. This opens up possibilities for real-world applications, such as high-speed optical flow in drone navigation or glare-free object detection in augmented reality devices.

Lastly, the combination of CI and CV systems reveals a pathway toward adaptive imaging pipelines that are task-aware and scene-aware. Unlike traditional imaging, where the captured data is static and uniform, computational imaging allows dynamic control over data acquisition. For instance, in a robotic vision system [64], light field cameras can adaptively focus on regions of interest for depth estimation, while high-speed imaging sensors can prioritize fast-moving targets for motion analysis. This task-aware imaging pipeline ensures that the input data is optimized for specific CV tasks, improving both efficiency and accuracy.

## V. FUTURE DIRECTIONS AND CHALLENGES

One promising direction is the development of task-specific imaging systems that dynamically adapt to the requirements of particular CV applications. Unlike conventional static imaging systems, future CI systems could use real-time feedback from CV algorithms to optimize the image acquisition process. For example, light field cameras could adjust their angular sampling based on the depth complexity of a scene, or high-speed cameras could focus on regions with rapid motion to prioritize temporal resolution. Such task-aware systems would maximize efficiency and ensure that the captured data aligns directly with the needs of the downstream CV tasks.

Another key direction involves the use of deep learning and data-driven methods to further enhance computational imaging pipelines. While traditional CI techniques rely on physical models and optimization methods, neural networks have shown remarkable capabilities in learning end-to-end mappings between raw sensor data and task-specific outputs. For instance, deep learning-based HDR imaging algorithms could reduce motion artifacts and improve real-time performance, while neural networks for light field imaging might simplify depth reconstruction by learning implicit geometric priors. Bridging deep learning with CI opens up opportunities for faster, more accurate, and more flexible imaging solutions. The integration of multi-modal imaging systems represents another important avenue for future exploration. Combining spatial, angular, spectral, and temporal imaging methods could lead to richer representations of scenes, enabling CV systems to perform robustly in highly complex environments. For example, integrating spectral imaging with light field techniques could improve material classification tasks, while combining high-speed imaging with HDR imaging could enable detailed motion analysis under challenging lighting conditions. Multi-modal imaging pipelines would allow CV systems to exploit complementary sources of information, resulting in superior performance across diverse tasks.

Despite the promising opportunities, several challenges remain in the integration of CI and CV. One major issue is the computational complexity of CI methods, particularly for real-time applications. Techniques like light field imaging and multi-exposure HDR often involve large volumes of data that require significant processing power. Developing efficient algorithms and hardware accelerators, such as GPUs and dedicated neural processors, will be crucial for enabling the deployment of CI-enhanced CV systems in resource-constrained environments like edge devices and mobile platforms.

Another challenge lies in the trade-offs between hardware and software. While CI techniques rely on specialized optical designs or sensor architectures, these systems must remain practical, affordable, and scalable for widespread adoption. Balancing the physical complexity of imaging systems with computational post-processing remains a key hurdle. Moreover, ensuring interoperability between emerging CI hardware and existing CV frameworks requires standardized pipelines and datasets that facilitate seamless integration.

The issue of data quality and generalizability also poses challenges. While CI techniques improve data fidelity, their performance is still dependent on specific imaging conditions. For example, light field cameras may struggle with sparse angular sampling in real-time scenarios, while HDR imaging can introduce artifacts in scenes with rapid motion. Addressing these limitations will require more robust algorithms capable of handling imperfect or incomplete input data across diverse operating conditions.

The convergence of computational imaging and computer vision presents unique opportunities for transforming various real-world applications. In autonomous vehicles, adaptive imaging systems combining HDR and light field techniques could provide unparalleled robustness to lighting variations and depth ambiguities. In surveillance and security, CI-enhanced face recognition systems capable of handling occlusions, glare, and extreme poses will significantly improve identification accuracy. Similarly, in robotics and agriculture, task-aware imaging pipelines can optimize performance for navigation, object recognition, and resource monitoring.

Furthermore, emerging areas such as augmented reality (AR) and virtual reality (VR) will greatly benefit from CI-integrated CV systems. High-resolution light field imaging, combined with HDR and deblurring techniques, can deliver immersive, artifact-free experiences by accurately reconstructing depth and preserving visual fidelity. In medical imaging, CI methods have the potential to revolutionize diagnostics and minimally invasive surgeries by providing high-precision, depth-aware representations of complex anatomical structures.

Finally, the rise of edge computing and AI accelerators presents an opportunity to deploy CI-CV systems in real-time applications. Advances in lightweight neural networks and ef- ficient hardware architectures will enable real-time processing of computationally intensive CI data on edge devices, such as autonomous drones, mobile phones, and IoT sensors [65].

## VI. CONCLUSION

This paper explored the synergies between computational imaging (CI) techniques and computer vision (CV) applications, showcasing how advanced imaging methods improve the robustness, accuracy, and versatility of modern CV systems. While CI has significantly enhanced tasks such as object detection, depth estimation, face recognition, and optical flow computation, the integration of these two domains is still evolving, offering a wealth of opportunities for further research and development.